%% file: main.tex
\documentclass[journal]{IEEEtran}
\usepackage{cite}
\usepackage{amsmath,amssymb,amsfonts}
\usepackage{algorithmic}
\usepackage{listings}
\usepackage[table]{xcolor}
\usepackage{multirow}
\usepackage{booktabs}
\usepackage[hidelinks]{hyperref}
\usepackage{cleveref}
\usepackage{algorithm}
\usepackage{comment}
\crefname{figure}{Fig.}{Figs.}
\crefname{table}{Table}{Tables}
\crefname{algorithm}{Algorithm}{Algorithms}
\crefname{section}{Section}{Sections}
\usepackage{graphicx,color}
\usepackage{subcaption}

\def\BibTeX{{\rm B\kern-.05em{\sc i\kern-.025em b}\kern-.08em
    T\kern-.1667em\lower.7ex\hbox{E}\kern-.125emX}}
\begin{document}

\title{Unsupervised Head Network Personalization for Split Computing via Federated Learning}

\author{Yutaro~Horikawa and Takayuki~Nishio%
\thanks{This work has been submitted to the IEEE for possible publication. Copyright may be transferred without notice, after which this version may no longer be accessible.}%
\thanks{Y. Horikawa and T. Nishio are with the School of Engineering, Institute of Science Tokyo, Tokyo 152-8550, Japan (e-mail: nishio@ict.eng.isct.ac.jp).}%
\thanks{This work was supported in part by JST PRESTO Grant Number JPMJPR2035, and JST ASPIRE Grant Number JPMJAP2346, Japan.}}

\markboth{}{Horikawa and Nishio: Unsupervised Head Network Personalization for Split Computing via Federated Learning}

\maketitle

\input{0_abstract}

\begin{IEEEkeywords}
Distributed inference, split computing, knowledge distillation, non-IID data, federated learning
\end{IEEEkeywords}

\input{1_introduction}
\input{2_related_works}

\input{preliminaries}
\input{3_proposed_method}

\input{4_evaluation}
\input{5_conclusion}

\bibliographystyle{IEEEtran}
\bibliography{main}

\end{document}

%% file: 0_abstract.tex
\begin{abstract}
Split Computing (SC) is a distributed inference method designed for balancing computational load and reducing latency by partitioning a neural network into head and tail networks and deploying them across both a mobile device and a server, respectively. Due to the limited bandwidth and computational power of mobile devices, the head network is often replaced with a neural network that requires lower computational and communication resources. However, it is challenging to prepare a sufficient quantity and quality of labeled data on the mobile device side for training the replacement head network. Furthermore, the data prepared by mobile users often contains sensitive privacy information, making data aggregation difficult.
In this paper, we propose a data-efficient head network training method for SC. This method utilizes federated learning and knowledge distillation to train the head network using a small amount of non-IID and unlabeled data from mobile devices, ensuring that the training datasets are not exposed. Moreover, by introducing personalization, we enhance the accuracy to levels comparable to or exceeding those of existing methods.
The experimental evaluations demonstrate that, despite data constraints imposed only on our method, it achieves accuracy comparable to the baselines. Furthermore, it significantly reduces the communication requirements by 95\% on mobile devices, with only a 3-5 percentage point drop in accuracy from the original model.
\end{abstract}

%% file: 1_introduction.tex
\section{Introduction}\label{sec:intro}

Deep learning and deep neural networks (DNNs) are critical in a wide range of applications \cite{hinton2015deep}. Furthermore, advancements in foundational models have enabled sophisticated tasks such as high-quality image generation \cite{dhariwal2021diffusion}, machine translation \cite{devlin2018bert}, and robotic control \cite{Vemprala24}. However, these advanced DNNs demand significant computational and memory resources, which challenges local computing. At the same time, the growing necessity to localize complex tasks such as speech recognition, personal assistants, and healthcare support \cite{biswas2023role} on mobile and edge devices for latency and privacy reasons makes it essential to adapt DNNs for these platforms. Yet, the inherently limited computational and memory capacities of mobile devices pose challenges for their real-time operation \cite{guo2018mobileinfer}.

Split computing (SC) has been proposed as an intermediate solution between cloud/edge server execution and local execution \cite{kang2017sc}. SC enables local devices to offload heavy computations of the DNN models to the cloud or edge servers. SC involves splitting the DNN into two parts: the first half processes on the device, which then sends the output of the middle layer to the server, and the server executes the remaining half. The first half of the split model is referred to as the Head Network, and the second half as the Tail Network. SC aims to reduce overall inference time and offers the added benefit of enhancing data privacy, as raw data is not transmitted to the server \cite{matsubara2021survey}.

In SC, it is often necessary to train a suitable dedicated model based on the constraints on computational performance and communication bandwidth of the local device, and replace it as the Head Network \cite{eshratifar2019bottlenet}. This is because general DNNs frequently map features to high-dimensional spaces near the input layer, resulting in very large intermediate representations. Using these as the Head Network leads to high memory consumption and communication overhead. Therefore, in SC, a layer with a smaller dimensionality for intermediate outputs, known as a bottleneck layer, is often used. By training a neural network with such a bottleneck layer, where the number of units and filters is limited, and replacing the original Head Network, the computational load and communication volume on the local device can be significantly reduced. 

Various methods for training Head Networks have been explored \cite{matsubara2019distill, matsubara2020hnd, matsubara2021ghnd, choi2020baf, sbai2021cde, yao2020dco}. However, these methods often assume the availability of sufficiently large datasets, particularly labeled training sets for supervised learning. In models using data related to images, audio, or healthcare, training data is not always accessible. Consequently, while local devices could potentially train their dedicated Head Networks using their own datasets, these datasets are often unlabeled, small in volume, and biased in distribution, making it difficult to train high-accuracy models.

To address the challenges arising from dataset constraints on local devices, this study proposes a new training method for SC which incorporates federated learning (FL) \cite{mcmahan2017fa} and head network distillation \cite{matsubara2019distill}. This method enables cooperative training of the head network across multiple local devices using a small amount of unlabeled and biased data from mobile devices, ensuring that the training datasets are not exposed. FL is a distributed model training framework that allows learning while keeping private data local. Additionally, the proposed method employs head network distillation, a knowledge distillation-based technique that can train on unlabeled data. By combining FL with personalization, it can train under severe data constraints without significantly degrading model performance. This approach is capable of training smaller models that decrease the overall inference time on SC, with only a minimal loss in accuracy, while preserving privacy under the stringent data constraints of the involved devices. The contributions of this paper are summarized as follows:

\vspace{2mm}
\noindent \textbf{A head network training method using small, unlabeled, and biased local datasets while preserving data privacy}: 
    This contribution outlines the development of a novel training method for the head network in SC that utilizes federated learning and head network distillation. The method ensures data privacy by keeping training datasets local and addresses the challenge of using small, unlabeled, and biased data from mobile devices.

\vspace{2mm}
\noindent \textbf{Performance evaluation using benchmark datasets}: We conduct a thorough performance evaluation of the proposed method using CIFAR-10 and CIFAR-100 under a non-IID setting, comparing it with baseline methods. The method reduces communication traffic by 95\% on local devices while maintaining accuracy within 3-5 percentage points of the original model. The personalization we introduce significantly enhances accuracy, achieving results comparable to the original model despite stringent data constraints.

\vspace{2mm}
The rest of the paper is organized as follows. We discuss related works in \cref{sec:related_works}. \cref{sec:preliminaries} introduces the building blocks of our method: split computing, knowledge distillation, and federated learning. \cref{sec:proposed_method} describes the proposed training method. \cref{sec:eval} provides the experimental evaluation. Finally, \cref{sec:conclusion} concludes this paper.

%% file: 2_related_works.tex
\section{Related Works}\label{sec:related_works}
This section reviews split computing and existing approaches to training models suitable for split computing, and clarifies the position of our contribution.

\subsection{Split Computing}\label{subsec:splitcomputing}
SC is a technique that divides the DNN into two parts and distributes the inference process between mobile devices and servers \cite{kang2017sc}. 
A trained model is divided into a head network deployed on a mobile client and a tail network deployed on a server; the client transmits the intermediate output of the head network to the server, which completes the inference. The formal notation of this procedure, used throughout this paper, is introduced in \cref{subsec:sc_formulation}.

SC can reduce computational latency by distributing the neural network's processing between the local device and the server. However, care must be taken regarding the communication latency associated with transmitting the intermediate output. Many neural networks map input features to a higher-dimensional space in layers close to the input layer. Thus, simply splitting an existing model can increase communication latency, potentially raising the total inference latency.
Therefore, many existing studies propose replacing the original model with models designed to be more suitable for split computing. Specifically, these models are designed to map to a low-dimensional space, referred to as a bottleneck, at the split point \cite{matsubara2019distill, eshratifar2019bottlenet}.

\subsection{Training Methods for Split Computing Models}\label{subsec:scmethods}

\begin{table*}[!t]
  \centering
  \caption{Summary of related works on head network training.\\
  A: model accuracy, C: computation load, D: amount of data transmitted, L: latency, CE: cross-entropy, KD: knowledge distillation, reconst.: reconstruction loss}{
  \resizebox{0.8\textwidth}{!}{
    {\renewcommand{\arraystretch}{0.9}
    \begin{tabular}{c c c c c c}
      \toprule
      \multirow{2}{*}{} & \multirow{2}{*}{} & \multirow{2}{*}{} & &Data constraints \\ \cline{4-6} 
  Research  &   Metrics  &  Loss   & Unlabeled & Decentralized & Non-IID  \\
      \midrule
      \cite{eshratifar2019bottlenet} & A, D, L & CE & -- & -- & -- \\
      \cite{hu2020fastcnn} & A, D, L & CE & -- & -- & --  \\
      \cite{matsubara2020hnd} & A, C, D, L & KD & $\checkmark$ & -- & -- \\
      \cite{matsubara2021ghnd} & A, C, D, L & KD & $\checkmark$ & -- & -- \\
      \cite{choi2020baf} & A, D & reconst. & $\checkmark$ & -- & -- \\
      \cite{sbai2021cde} & A, C, D & reconst. + KD & $\checkmark$ & -- & --\\
      \cite{yao2020dco} & A, D, L & reconst. + KD & $\checkmark$ & -- & --\\
      \cite{shimizu2022NASC} & A, L & CE + Latency loss & -- & -- & -- \\
      \cite{dong2022splitnets} & A, L & CE + Latency loss & -- & -- & --\\
      
      \begin{tabular}{c}
        \textbf{Ours}
      \end{tabular} & A, C, D & KD & $\checkmark$ & $\checkmark$ & $\checkmark$ \\
      \bottomrule
    \end{tabular}}\label{tab:related_works}
    }
}
\end{table*}

The methods for efficiently training models suitable for split computing are a critical issue in this field, and various approaches have been proposed. Two particularly important challenges are as follows:

\vspace{2mm}
\noindent \textbf{Designing model architectures appropriate for split computing}:
The first challenge is optimizing the model architecture, especially the head network and the split point, considering the limitations of the mobile device's computational power and communication bandwidth. Neural network architectures have nearly infinite combinations of layers, types, units, and filters, making it essential to determine an optimal architecture. This task is part of the domain known as Neural Architecture Search (NAS). Traditional NAS focuses on metrics such as accuracy, computational cost, and latency, with methods aimed at maximizing accuracy or improving the accuracy-computation trade-off while minimizing computation for searching the architecture and training the model parameters \cite{Akimoto19, hwnasbench}. However, for split computing, it is crucial to also consider communication cost and latency.

Kang et al. focus on identifying optimal split points that balance computation latency between the server and local devices, as well as communication latency. They conducted a comprehensive investigation on well-known benchmark models \cite{kang2017sc}. Drawing from these insights, they design a protocol to optimize the split point. Bakhtiarnia et al. propose dynamic split computing, which adjusts the splitting point based on varying communication bandwidth to minimize total latency \cite{DynamicSC}. These studies do not alter the structure of pre-trained models but only adjust the split points.

To improve the trade-off between accuracy and latency, particularly by reducing communication traffic, a widely used approach is to insert a layer with a bottleneck structure, which has a lower dimensionality than the original structure, at the split point. This approach is akin to using an incomplete autoencoder \cite{choi2020baf, sbai2021cde, yao2020dco}. However, in many studies, the inserted bottleneck structures are designed manually or empirically. On the other hand, methods based on the NAS framework have also been proposed to explore structures suitable for split computing \cite{shimizu2022NASC, dong2022splitnets}. By incorporating model accuracy and total latency, including both computation and communication, into the objective function or constraints of NAS, it is possible to search for the optimal architecture from among the candidates. In particular, our previous research \cite{shimizu2022NASC} propose a method to simultaneously search for model structures and parameters using the Adaptive Stochastic Natural Gradient method. However, these studies require substantial computational resources and training data for structure search, model parameter training, and fine-tuning.

\vspace{2mm}
\noindent \textbf{Efficient Training Methods for Modified Models}:
Efficiently training the parameters of models updated to have structures suitable for split computing is a significant practical challenge. In environments where split computing is applied, access to abundant computational resources and training data is not always guaranteed. Therefore, it is essential to train models that achieve accuracy comparable to the original models with minimal computational effort, memory consumption, and training data.

When ample labeled training data and computational resources are available, the most straightforward training method is to perform end-to-end training of the entire model using backpropagation based on loss functions such as cross-entropy (CE), similar to standard model training practices \cite{eshratifar2019bottlenet, hu2020fastcnn}. However, sufficient labeled data is not always accessible. Particularly when leveraging publicly available open models, the data used for their training may not be available.

In cases where a bottleneck layer resembling an incomplete autoencoder is inserted, training methods that utilize reconstruction errors have been proposed. These methods do not require labeled data for training \cite{choi2020baf, sbai2021cde, yao2020dco}. By training the newly inserted autoencoder section to match its input and output using unlabeled data, the signal changes before and after the bottleneck are minimized, maintaining inference accuracy. However, this approach is limited to autoencoders, and adding new layers increases total computation. Thus, it is challenging to apply this method when reducing computational load on the local device or server is prioritized, even if it means sacrificing some accuracy.

To address this, a training method using knowledge distillation (KD), so called Head network distillation (HND), has been proposed \cite{matsubara2020hnd, matsubara2021ghnd}. 
KD \cite{hinton2015distill} is introduced to efficiently train smaller models (called student models) to achieve comparable accuracy to larger pre-trained models (called teacher models). 
Matsubara et al. adapt KD for training models in split computing, particularly for the head network. They propose a method that efficiently trains only the head network (specifically, the head network and decoder part) using a portion of the original model \cite{matsubara2020hnd}. This method achieves higher accuracy using only unlabeled data compared to end-to-end training with cross-entropy loss on labeled datasets and conventional KD-based training that involves entire models. However, it still requires a substantial amount of centralized unlabeled dataset.

The integration of FL and KD itself has been explored since the early days of FL. Federated distillation, pioneered by Jeong et al.~\cite{jeong2018fd}, exchanges model outputs (logits) instead of model parameters to drastically reduce the training-time communication of FL under non-IID data, and Itahara et al.~\cite{itahara2023distillation} extended this line to semi-supervised settings in which an unlabeled shared dataset is exploited through distillation for communication-efficient collaborative training. Subsequent works follow this direction: FedKD~\cite{wu2022fedkd} mutually distills knowledge between a local teacher and a globally shared student model, and Li et al.~\cite{li2023uavkd} combine KD with transfer learning for communication-efficient FL in UAV networks. These studies employ KD to reduce the \emph{training-time} communication of FL itself, and their goal is to obtain a full inference model on each client. In contrast, our method employs distillation for a fundamentally different purpose: it transfers the intermediate representations of a pre-trained model to a lightweight head network so that the \emph{inference-time} communication and computation of SC are reduced, and the distillation target itself eliminates the need for labels on the client side. Recent studies on learned feature compression for SC, such as FrankenSplit~\cite{furutanpey2024frankensplit}, also reduce inference-time communication, but assume that training data can be centralized. To the best of our knowledge, no prior work trains SC head networks under the combination of unlabeled, decentralized, and non-IID data constraints, as summarized in \cref{tab:related_works}.

Inspired by HND, this study proposes an efficient head network training method under data constraints, particularly when the data is unlabeled and distributed across local devices, with each device having access to a limited amount of data. \cref{tab:related_works} summarizes existing research and our work, highlighting the differences among them. The proposed method overcomes data constraint challenges by integrating federated learning and HND. Additionally, the method mitigates accuracy degradation caused by non-IID data introduced by federated learning through a model personalization mechanism. This approach achieves accuracy comparable to original and centrally trained models, even when the data distribution varies significantly across local devices.

%% file: preliminaries.tex
\section{Preliminaries}\label{sec:preliminaries}
This section introduces the building blocks of the proposed method: the formulation of split computing, knowledge distillation and head network distillation, and a brief overview of federated learning.

\subsection{Split Computing Formulation}\label{subsec:sc_formulation}
We consider a neural network model $f(x;w)$ parameterized by $w$. For example, in the case of a 3-layer neural network, the model can be represented as
\begin{equation}\label{eq:nn}
    f(x;w) = l_3(l_2(l_1(x))),
\end{equation}
where $x$ is the input and $l_i$ is a mapping of the $i$-th layer parameterized by $w_i$. In this manner, the computation of a neural network can be regarded as an operation where the input sequentially passes through multiple functions. Thus, we can divide the model into two composite functions, such as $l_1$ and $l_3 \circ l_2$. The first part is called the head network, $f_\mathrm{head}= l_1$, and is deployed to a mobile client, while the second part is called the tail network, $f_\mathrm{tail} = l_3 \circ l_2$, and is deployed to a server.

In the distributed inference procedure, the mobile client feeds input $x$ and obtains the intermediate output of the head network,
\begin{equation}\label{eq:nn_head}
    z = f_\mathrm{head}(x) = l_1(x).
\end{equation}
The intermediate output $z$ is transferred to the server and used as the input for the tail network, resulting in the inference output
\begin{equation}\label{eq:nn_tail}
    y = f_\mathrm{tail}(z) = l_3(l_2(z)).
\end{equation}

\begin{figure}[!t]
  \centering
  \begin{minipage}[b]{0.9\columnwidth}
    \centering
    \includegraphics[width=0.85\columnwidth]{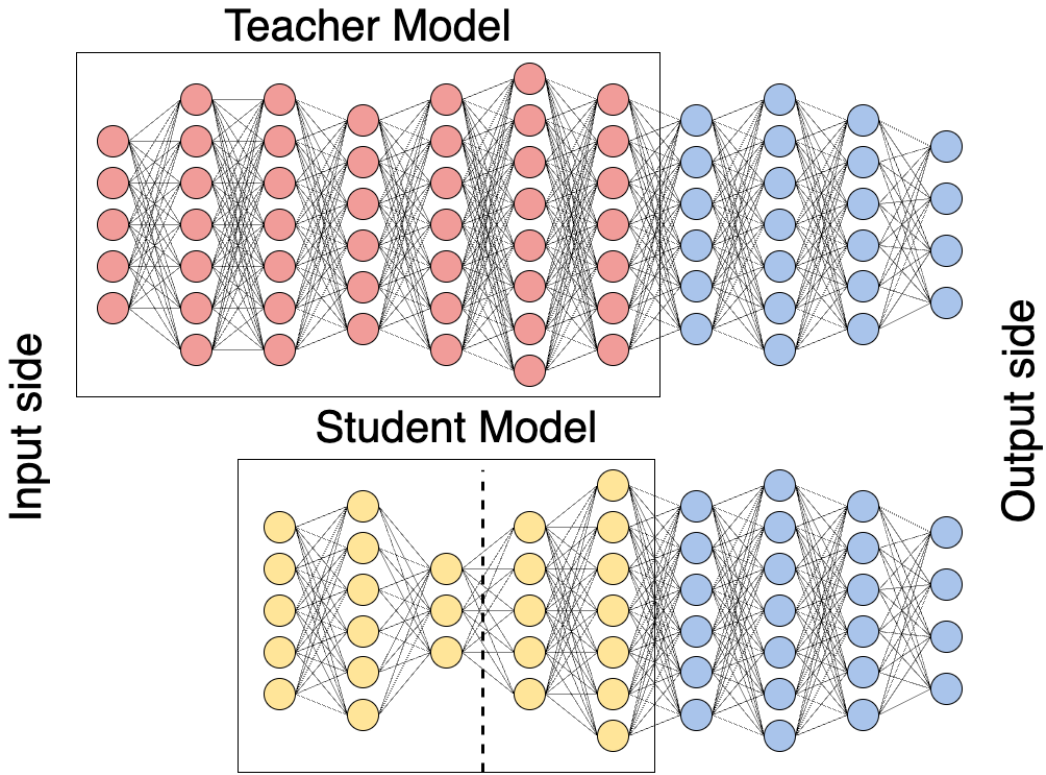}
    \subcaption{Teacher and student models}
    \label{fig:hndmodels}
  \end{minipage}\\[2mm]
  \begin{minipage}[b]{0.9\columnwidth}
    \centering
    \includegraphics[width=0.95\columnwidth]{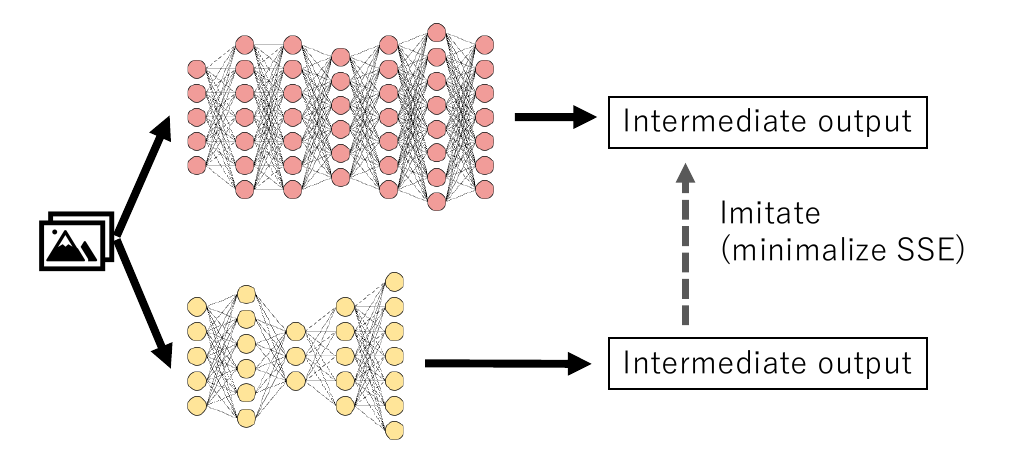}
    \subcaption{Training procedure}
    \label{fig:hndtrain}
  \end{minipage}
  \caption{Head Network Distillation~\cite{matsubara2020hnd}: (a) teacher and student models extracted from the original pre-trained model, and (b) the training procedure minimizing the error between the intermediate outputs.}
  \label{fig:hnd}
\end{figure}

\subsection{Knowledge Distillation and Head Network Distillation}\label{subsec:hnd}
HND \cite{matsubara2020hnd} is a training method for the head network based on knowledge distillation \cite{hinton2015distill}. Knowledge distillation leverages a well-trained large model, called the teacher model, to train an untrained small model, called the student model. In knowledge distillation, the student model is trained using the output of the teacher model, referred to as a soft target, as well as the labels of data samples, referred to as hard targets. This allows the student model to achieve accuracy comparable to the teacher model while maintaining a smaller model size. HND applies knowledge distillation to the intermediate outputs of the model, enabling the training of only a part of the model using unlabeled data.

\cref{fig:hndmodels} shows an example of HND's teacher and student models extracted from the original pre-trained model. Unlike conventional knowledge distillation, which uses the entire model, HND uses only several layers from the input side of the model as the teacher model. The number of layers used is a hyperparameter. A student model, with the same input and output layers as the teacher model but with fewer parameters and a bottleneck structure, is prepared. This bottleneck part, indicated by the dashed line in \cref{fig:hndmodels}, divides the model. The input-side portion is used as the head network, while the remaining part, combined with the rest of the original model excluding the teacher model, forms the tail network used in split computing. In HND, the student model is trained to minimize the errors between the outputs of the teacher and student models, ensuring consistency, as shown in \cref{fig:hndtrain}. Specifically, the loss function $L$ is defined as follows:
\begin{equation}\label{eq:hndloss}
    L = \sum_{x\in X} \|f_\mathrm{T}(x;\, w_\mathrm{T}) - f_\mathrm{S}(x;\, w_\mathrm{S})\|^2,
\end{equation}
where $X$ is a set of training data, $x$ is a data sample, and $f_\mathrm{T}(x;\, w_\mathrm{T})$ and $f_\mathrm{S}(x;\, w_\mathrm{S})$ are the outputs of the teacher model parameterized by $w_\mathrm{T}$ and student model parameterized by $w_\mathrm{S}$, respectively, with $x$ as input. By training the student model, its output becomes identical or very close to the output of the teacher model. Consequently, the processing results in the subsequent layers also yield the same outcomes as when using the teacher model. This ensures that the inference results remain consistent whether the original model is used or the teacher model is replaced with the student model.

\subsection{Federated Learning}\label{subsec:fl_prelim}
FL \cite{mcmahan2017fa} is a distributed training framework in which multiple clients collaboratively train a shared model while keeping their data local. In the representative FedAvg algorithm \cite{mcmahan2017fa}, a server distributes a global model to the clients; each selected client updates the model locally on its own dataset for several epochs; and the server aggregates the returned models, typically by weight averaging, to update the global model. This procedure is iterated over multiple rounds until convergence. Since only model parameters are exchanged, FL preserves the privacy of the clients' raw data. A well-known challenge of FL is the accuracy degradation under non-IID client data, which has motivated a large body of work on personalized federated learning \cite{matsuda2024pflbenchmark, fallah2020pfl, shamsian2021pfedhn}.

%% file: 3_proposed_method.tex
\section{Federated Head Network Personalization}\label{sec:proposed_method}
This section presents the proposed method, FedHNP. We first describe the system model and assumptions, then provide an overview and the detailed two-stage procedure, and finally illustrate the model architecture with a concrete example.

\begin{figure}[!t]
  \centering
  \includegraphics[width=0.45\textwidth]{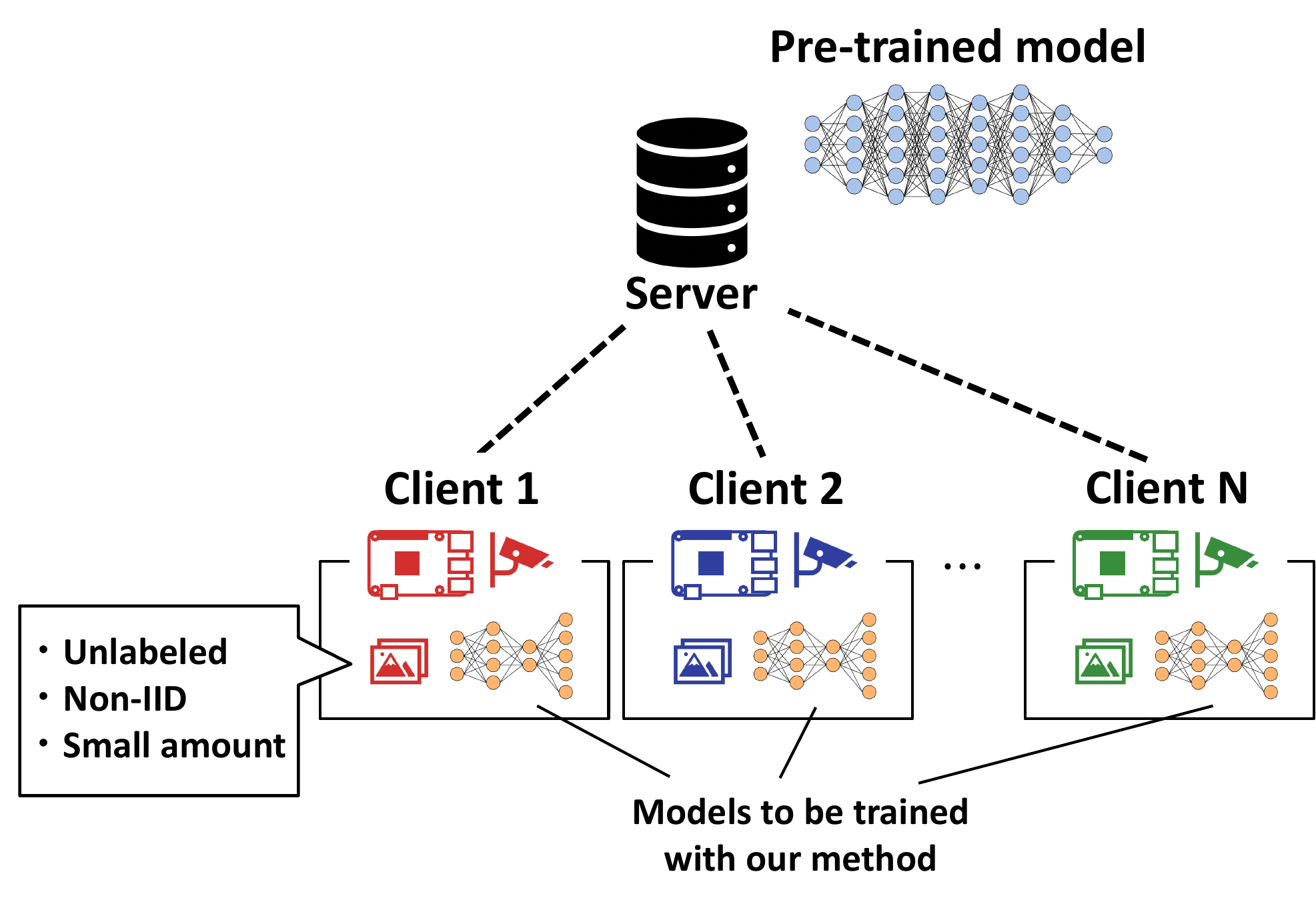}
  \caption{System model}\label{fig:systemmodel}
\end{figure}

\subsection{System Model}\label{subsec:systemmodel}
\cref{fig:systemmodel} shows the system model. The system consists of a server and $N$ edge devices such as Raspberry Pi and Jetson Nanos. Following the terminology of federated learning, these edge devices are referred to as clients in this paper.

First, we outline the assumptions regarding communication. Each client communicates only with the server during training with FL and distributed inference with SC, but not directly with other clients. There are bandwidth limitations on communication between the client and the server, and it is necessary to reduce the amount of data sent to the server during machine learning inferences to minimize execution time. On the other hand, we assume that model training is performed independently of inference processing, during times with available bandwidth, such as overnight, making the reduction of communication volume during training outside the scope of this study.

Next, we outline the assumptions regarding computational capability. The clients lack the computational capability to efficiently process the server's high-performance pre-trained models within the desired latency for real-time tasks. In contrast, these devices can swiftly handle smaller models. Additionally, the urgency for rapid processing is less critical during the training phase than during inference, as real-time performance is not required for model training, making speed less of a priority.

Then, we outline the assumptions regarding training data. Each client collects unlabeled data samples, which are used for training and inference. These clients solve common tasks, such as object detection or anomaly detection, but the data sample distributions observed in each client's environment differ, resulting in a non-IID setting when the clients' data is aggregated. Additionally, the amount of data available to each client is limited, making it difficult to train a high-accuracy model independently, and this data cannot be collected or aggregated on the server due to privacy concerns.

Finally, we outline the assumptions regarding the machine learning model. The server has a pre-trained model, assumed to be well-trained with labeled data, but the labeled training dataset is unavailable. This is not an unusual assumption, as it often occurs when using publicly available models. For this study, we focus on the training process, assuming that the structures of the original model and the head and tail networks are predetermined and known to the server and each client. The head network is designed to be small enough to be trained and processed by the client, while the optimization of the model's structure is outside the scope of this study.

\subsection{Overview of Proposed Method}\label{subsec:overview}
\begin{figure*}[!t]
  \centering
  \includegraphics[width=0.95\textwidth]{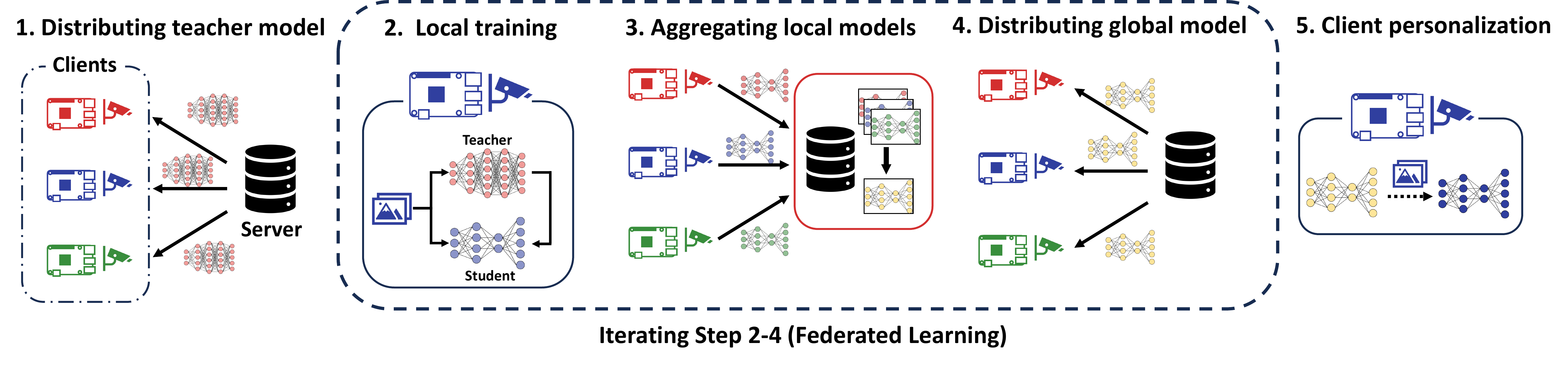}
  \caption{Proposed method: Federated Head Network Personalization}\label{fig:method}
\end{figure*}

This section provides an overview of our proposed FedHNP method, as shown in \cref{fig:method}. FedHNP enables the training of models more suitable for SC, with reduced computation and communication costs, while maintaining the accuracy of the original model. The algorithm of the proposed FedHNP is shown in \cref{alg:step1}. The training method is divided into two stages: FL-based global model training and client personalization.

\begin{algorithm}[!t]
    \caption{Federated Head Network Personalization}
    \label{alg:step1}
    \begin{algorithmic}[0]
    \STATE \textbf{Stage 1. FL-based global model training}
    \STATE \textbf{1. Initialization} \\ 
        Obtain teacher model from pre-trained model, \\
        and distribute it to each client \\
        Initialize the parameter of all client models $w_k$ and\\ 
        the global model $w_\mathrm{g}$
    \STATE \textbf{2. Local update} \\
        Select $C$ clients out of $N$\\
        \FOR{Each selected client $k$ \textbf{in parallel}}
            \FOR{$e=1,\dots, E_{\mathrm{FL}}$}
            \STATE Update each model $w_k$ locally
            \STATE $w_k \leftarrow w_k - \eta \nabla L_k$
            \ENDFOR
        \ENDFOR
    \STATE \textbf{3. Aggregation} \\
        \STATE $C$ clients upload their models $\hat{w}_k$ to the server\\
        Aggregate the models on the server and update\\ 
        the global model
        \STATE $w_g \leftarrow \displaystyle\frac{1}{C}\sum_{i\in I_C} \hat{w}_i$
    \STATE \textbf{4. Distribution} \\
        Distribute the global model to all clients
    \STATE 
    \STATE Steps 2-4 are iterated for multiple rounds $R$
    \STATE
    \STATE \textbf{Stage 2. Client personalization}
    \FOR{Each client $k$ \textbf{in parallel}}
        \FOR{$e=1,\dots, E_{\mathrm{p}}$}
            \STATE Update each model $w_k$ locally
            \STATE $w_k \leftarrow w_k - \eta \nabla L_k$
        \ENDFOR
    \ENDFOR
    \end{algorithmic}
\end{algorithm}

The FL-based global model training stage aims to train a base common model, referred to as the global model, using the FL procedure. By introducing FL, each client can contribute to training the global model without aggregating data, thus preserving the privacy of the clients' data, even when each client has only a small amount of data.

The training flow follows the well-known FedAvg algorithm \cite{mcmahan2017fa}. In the original FedAvg, an initialized global model is shared with all the clients, who then train the model locally using conventional gradient descent methods with output loss. The locally trained models, called client models, are then gathered and aggregated to update the global model. In contrast, in the proposed FedHNP, to enable training with unlabeled data, the local training on the client side is replaced with Head Network Distillation (HND) \cite{matsubara2020hnd}. The details of HND are described in \cref{subsec:hnd}. HND allows for unsupervised training using unlabeled data by ensuring that the intermediate outputs of the trained model and the target model are consistent.

The second stage is the personalization of the well-trained models to fit the clients' local non-IID data. In this paper, we employ a fine-tuning method for personalization, which is simple and straightforward yet effective for many tasks \cite{matsuda2024pflbenchmark}. This personalization can significantly improve the performance of client models against each client's local data.

\subsection{Details of FedHNP}\label{subsec:fedhnp}

This section describes our proposed FedHNP in detail. FedHNP integrates distillation-based model training based on HND and cooperative training based on FL and incorporates a personalization method. This is the key difference between FedHNP and original HND. As briefly explained in the previous section, FedHNP consists of the following two stages.

\subsubsection{FL-based Global Model Training}\label{subsubsec:step1}

A global model is trained using the distributed clients' local dataset based on the FL algorithm. \cref{alg:step1} describes the detailed procedures.

\noindent\textbf{1. Initialization.}\quad
First, the server determines which layers of the original model will constitute the teacher model and extracts the teacher model from the pre-trained original model for HND. The server also determines the architecture of a student model considering the computation and communication capabilities of the local devices, initializes the parameters of the student models $w_k$ (where $k=1,2,\dots,N$, indicating the identification of a client), and distributes the teacher model and a student model to each client. The student models, which are stored and updated by the clients, are referred to as client models. The architectures of the teacher and student models are hyperparameters. Although the methods to optimize them are beyond the scope of this paper, we discuss the model architecture design with specific examples in Section \ref{subsubsec:example}.

\vspace{2mm}
\noindent\textbf{2. Local update.}\quad 

From the $N$ clients, $C$ clients are randomly selected. These selected clients update their client models and upload them to the server, thereby reducing communication traffic and preventing congestion in the server networks.%

Each selected client $k$ trains its client model $w_k$ by HND's algorithm using only its dataset $X_k$. 
Redefining the loss function $L_k$ for client $k$ using \cref{eq:hndloss}, we obtain
\begin{equation}
    L_k = \sum_{x\in X_k} \|f_\mathrm{T}(x;\, w_\mathrm{T}) - f_\mathrm{S}(x;\, w_k)\|^2,
\end{equation}
and the model parameter is updated via the stochastic gradient descent method as follows:
\begin{equation}\label{eq:update}
    w_k \leftarrow w_k - \eta \nabla L_k,
\end{equation}
where $\eta$ is the learning rate.
Clients update their client models for a predetermined number of epochs $E_{\mathrm{FL}}$. The updated client models $\hat{w}_k$ are sent to the server. 

\vspace{2mm}
\noindent\textbf{3. Aggregation.}\quad
The server aggregates the updated client models to create an updated global model with the parameter $w_\mathrm{g}$. Although there are various methods for aggregation, this paper employs a simple arithmetic mean, which is widely used in many FL algorithms, as represented below:
\begin{equation}\label{eq:aggregation}
    w_g \leftarrow \frac{1}{C}\sum_{i\in I_C} \hat{w}_i,
\end{equation}
where $I_C$ is the index set of the selected clients.

\vspace{2mm}
\noindent\textbf{4. Distribution.}\quad
The global model $w_\mathrm{g}$ is distributed to all clients. Clients then replace their client models with $w_k$ as follows:
\begin{equation}\label{eq:distribution}
    w_k \leftarrow w_g.
\end{equation}

The procedures from "2. local update" to "4. distribution" are repeated for a predefined number of iterations $R$ or until certain training and validation losses are achieved.

\subsubsection{Client Personalization}\label{subsubsec:step2}
The clients update their client models to better fit their local data distribution, thereby improving model accuracy for their specific data domains. The model personalization mechanism in federated learning, known as Personalized Federated Learning (PFL), has been widely studied to address the accuracy degradation caused by non-IID data~\cite{matsuda2024pflbenchmark}. Representative approaches include meta-learning-based methods that optimize the global model as an initialization for fast local adaptation~\cite{fallah2020pfl} and hypernetwork-based methods that generate client-specific parameters~\cite{shamsian2021pfedhn}. This paper adopts local fine-tuning for three reasons. First, a recent comprehensive benchmark reports that fine-tuning consistently yields stable accuracy across diverse tasks and often matches or outperforms more sophisticated PFL methods~\cite{matsuda2024pflbenchmark}. Second, fine-tuning is executed entirely on the client and requires no additional communication or server-side coordination, which suits the bandwidth-limited system model in \cref{subsec:systemmodel}. Third, unlike meta-learning or hypernetwork approaches, it introduces no additional models or hyperparameters to be maintained on resource-constrained clients. Fine-tuning directly addresses the non-IID challenge because the global model trained in Stage~1 provides a generalized initialization, while the subsequent local updates specialize the model to each client's own data distribution, which is consistent with the distribution the client encounters at inference time.

For personalization using fine-tuning, each client trains its client model, which is the well-trained global model, using only its own data through the same local update method as in Step 1. However, unlike during FL-based global model training, it is necessary to sufficiently fit the client model to the client's local data. Therefore, the number of fine-tuning epochs $E_p$ is set to be greater than $E_\mathrm{FL}$.

It is worth noting the overhead of the personalization stage. Since fine-tuning is performed locally, it incurs no communication cost; the total training-time communication of FedHNP is determined solely by the FL stage (per round, $C$ client-model uploads and the distribution of the global model, where the transferred student model is far smaller than the original model). The computational overhead of personalization is $E_p$ local epochs (10 for CIFAR-10 and 20 for CIFAR-100 in our experiments), which is marginal compared with the epochs consumed over the $R$ rounds of the FL stage. As shown later in \cref{fig:phase2} (\cref{sec:eval}), most of the accuracy gain is obtained within the first few epochs, indicating that the accuracy benefit of personalization is obtained at a small and controllable computational cost.

\subsubsection{Distributed Inference after Training}\label{subsubsec:inference}
During the inference phase, the client executes the head network, while the server executes the tail network. Therefore, it is necessary to transfer the remaining portion of the student model, after the head network, from the client to the server. As will be explained in more detail in the next section, this can be understood clearly by referring to the right side of \cref{fig:models}. The portion of the student model (shown in yellow) up to the splitting point, remains as the head network on the client, while the remaining part of the network is transferred to the server. The server combines this transferred network with the blue-colored portion, which was not used during training, to form the tail network. 

Since the fine-tuned student model parameters differ for each client, the tail network must be constructed individually for each client. Specifically, for client \( k \), the model is defined as:
\begin{equation}\label{eq:clientmodel}
    f_\mathrm{S}(\cdot;\, w_k) = f_\mathrm{D}(f_\mathrm{H}(\cdot;\, w^\mathrm{H}_k);\, w^\mathrm{D}_k),
\end{equation}
where \( f_\mathrm{H}(\cdot;\, w^\mathrm{H}_k) \) represents the head network of client \( k \) parameterized by \( w^\mathrm{H}_k \), and \( f_\mathrm{D}(\cdot;\, w^\mathrm{D}_k) \) represents the remaining part of the student model, parameterized by \( w^\mathrm{D}_k \), which is similar in structure to a decoder.
Let \( x \) be the input data, \( o_\mathrm{head} \) the output of the head model, and \( o_\mathrm{tail} \) the output of the tail model. These outputs are represented as follows:
\begin{align}\label{eq:output}
    o_\mathrm{head} &= f_\mathrm{H}(x;\, w^\mathrm{h}_k),\\
    o_\mathrm{tail} &= f_\mathrm{R}(f_\mathrm{D}(o_\mathrm{head};\, w^\mathrm{D}_k);\, w_\mathrm{R}),
\end{align}
where \( f_\mathrm{R}(\cdot;\, w_\mathrm{R}) \) is the remainder of the pre-trained model, excluding the teacher model, and is parameterized by the pre-trained weights \( w_\mathrm{R} \).
In distributed inference, the client first computes $o_\mathrm{head}$ and sends it to the server. The server then computes $o_\mathrm{tail}$ and that is the inference result.

\subsubsection{Model Architecture: Example Study}\label{subsubsec:example}
This section aims to clarify the relationships among the various models and sub-models discussed in this paper—namely, the Original model, Teacher model, Student model, head network, and tail network—using specific examples. Additionally, we will discuss the design guidelines for the student model.

\begin{figure}[!t]
  \centering
  \includegraphics[width=0.45\textwidth]{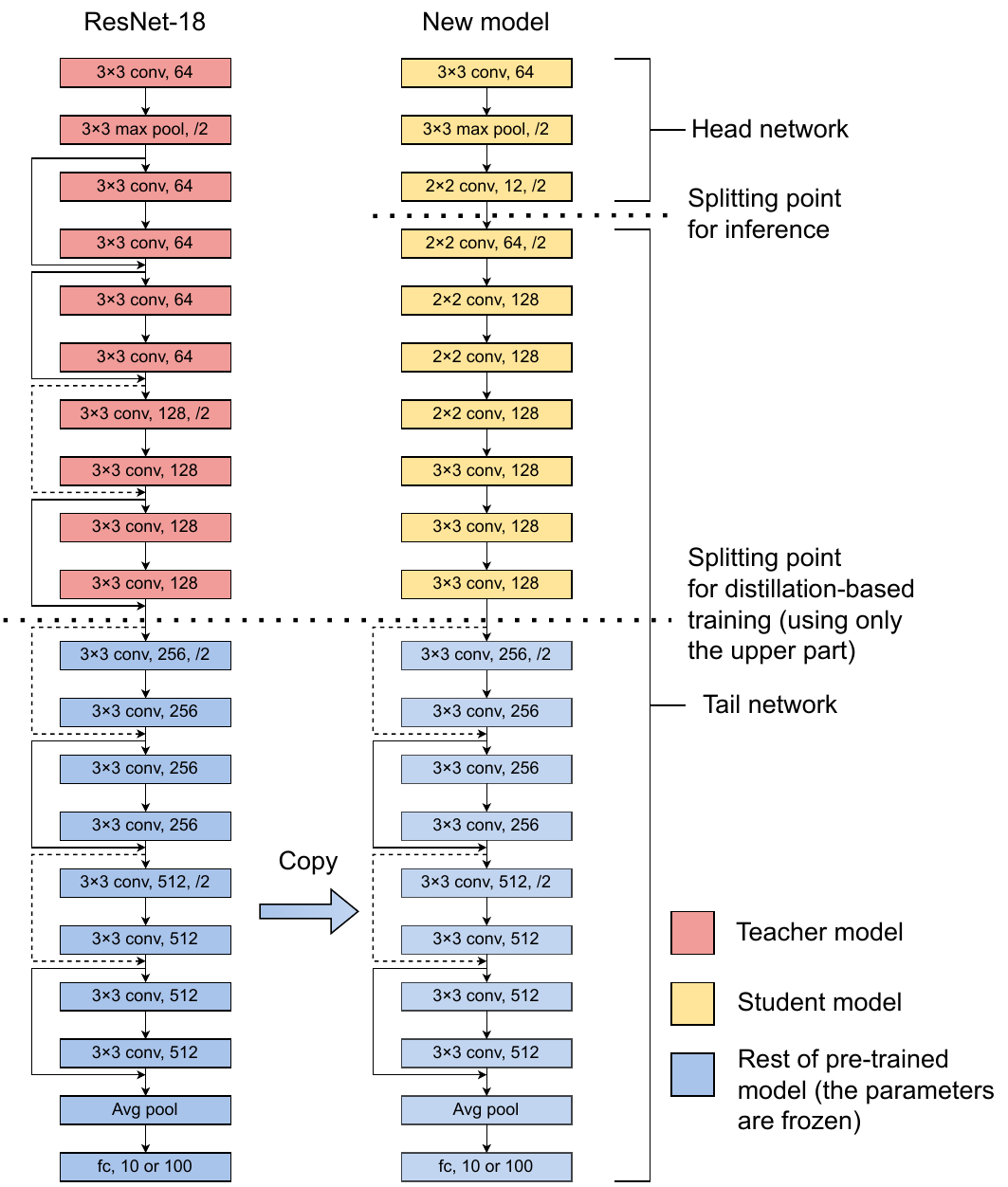}
  \caption{Architecture of the original and bottleneck injected models. Training uses only the teacher and student models.}\label{fig:models}
\end{figure}
\cref{fig:models} illustrates the model architecture used in our experiments as an example. We used ResNet-18 \cite{he2016resnet} as our original model. The left side of the figure presents a pre-trained model hosted on the server, while the right side shows a model that has been newly trained. During the model training phase, the red segment is used as the teacher model $f_\mathrm{T}(\cdot;\, w_\mathrm{T})$, and the yellow segment is updated as the student model $f_\mathrm{S}(x;\, w_\mathrm{S})$. The blue segment is $f_\mathrm{R}(\cdot;\, w_\mathrm{R})$, and its parameters $w_\mathrm{R}$, sourced from the original model, remain unchanged throughout the entire process. 

In the inference phase, the trained model on the right side is used. The initial segment of the yellow-colored head network constitutes the head network $f_\mathrm{H}(\cdot;\, w^\mathrm{h}_k)$, computed by client $k$. The subsequent portion is used as the tail network, $l_\mathrm{R} \circ l_\mathrm{D}$, and is processed by the server. Therefore, the output of the third convolution layer of the model on the right side corresponds to $o_\mathrm{head}$, and is transmitted via networks. The server receives $o_\mathrm{head}$ from the client and processes it through the tail network $l_\mathrm{R} \circ l_\mathrm{D}$.

Although the model shown in \cref{fig:models} is just an example, we will discuss key considerations when designing the architecture of the student model. Two particularly important aspects are:

\begin{itemize}
\item Model size: This refers to the number of model parameters. The computational time, particularly the size of the head network, significantly impacts processing time on the client.
\item Bottleneck size: This corresponds to the intermediate tensor size at the split point, i.e., the size of the output tensor from the head network. It greatly affects the amount of data transmitted during inference and, consequently, the communication latency.
\end{itemize}

While reducing both model size and bottleneck size can decrease computation and communication latency in SC, there is a trade-off with model accuracy. A model that is too small in size or has an excessively reduced bottleneck may significantly degrade inference accuracy, potentially failing to meet performance requirements.

In the model used for this study, we designed a head network with fewer parameters by reducing the filter size while largely preserving the structure of the original model. Specifically, the head network size was set to 5,080 parameters by reducing the number of filters from 64 to 12. This is significantly smaller than the 38,848 parameters that the original model would have had if split at the same point, resulting in a substantial reduction in computation latency on the client.

As for bottleneck size, the smaller it is, the smaller the intermediate outputs sent from the client to the server during each inference, which helps reduce communication latency. A simple strategy is to reduce the number of filters or units corresponding to the output layer of the head network, thus reducing the bottleneck size while keeping the original model’s structure. However, this strategy may not be applicable in certain cases, particularly in deep neural networks (DNNs) that employ skip connections, which add the input of one layer to the output of another. When the split point is set across a skip connection, the total tensor size increases because both the outputs and the input to the skip connection need to be transmitted. Therefore, it is important to avoid splitting at skip connections, though this may limit the available splitting points and make it more difficult to reduce the size of the head network.

To address these challenges, we designed a student model without skip connections and implemented a bottleneck with small filter sizes in the early layers. This design preserves much of the architecture of the original ResNet-18 model, except for modifications around the bottleneck and skip connections.

In our design, the bottleneck dimensionality is selected such that the transmitted tensor, $(12, 9, 9)$, is smaller than the raw input $(3, 32, 32)$, reducing the per-inference traffic to 1.0~KB with 8-bit quantization---one third of cloud offloading and $1/8$--$1/16$ of splitting the original model (\cref{tab:model_comp} in \cref{sec:eval})---while keeping the client-side computation comparable to the shallowest viable split point of the original model. In general, the bottleneck should be dimensioned as the smallest tensor for which the accuracy degradation remains within the application's tolerance; exploring this trade-off automatically, e.g., with NAS~\cite{shimizu2022NASC, dong2022splitnets}, is orthogonal to our training method.

It is worth reiterating that the student model presented here may not be optimal, as the ideal model configuration depends on factors such as communication speed, the computational resources available on the device, and the specific task being executed. Although some studies \cite{shimizu2022NASC, dong2022splitnets} have introduced NAS as a strategy for designing models tailored to SC, our study does not focus on optimizing the model architecture. Instead, our focus is on training methods for the head network under data constraints.

%% file: 4_evaluation.tex
\section{Evaluation}\label{sec:eval}
This section evaluates FedHNP on CIFAR-10 and CIFAR-100 under non-IID settings, in terms of computation and communication costs, accuracy against baselines, and the effect of personalization.

\subsection{Setup}\label{subsec:setup}
We use the CIFAR-10 and CIFAR-100 datasets \cite{krizhevsky2009cifar}, which are image datasets containing 10 and 100 different labels, respectively. Both datasets are pre-divided into training and test sets. We use 60\% of the training data to prepare a pre-trained original model, while the remaining 40\% is allocated to the client datasets. The dataset used for training the original model is assumed to be IID.

In contrast, we adopt a Non-IID setting for the client datasets in this study. Specifically, each client is assigned data for only 2 out of 10 labels in the CIFAR-10 dataset, and 10 out of 100 labels in the CIFAR-100 dataset. This type of Non-IID setting is also commonly used in related work \cite{shamsian2021pfedhn}.%

Specifically, we generate Non-IID client data in the following procedure. First, we allocate two or ten classes (for CIFAR-10/CIFAR-100) to each client. Then, for each client $i$ and its class $c$, we sample $\alpha_{i, c}\sim U(0.4, 0.6)$ and assign $\frac{\alpha_{i, c}}{\sum_j {\alpha_{j,c}}}$ of the class $c$ data. We do the same for $N$ clients and their classes. Of the allocated training data for each client, 20 percent is reserved for validation.

On the other hand, the test dataset is divided into $N$ portions, matching the number of clients. The distribution of the test dataset for each client is made consistent with the distribution of the client's training set, by using $c, \alpha_{i,c}$ assigned to the clients during the splitting of the training set.

\cref{fig:models} illustrates the structure of the teacher and student models. We use ResNet-18 \cite{he2016resnet} as the original model, which is pre-trained and owned by the server. The first half of the model is distilled and personalized to be more suitable for SC using the proposed method. During training, each client holds copies of the teacher model (which is pre-trained and frozen) and the student model, and the student model is updated locally.
In the inference phase, each personalized head network, extracted from the student model, is deployed on the client, while the corresponding tail networks are deployed on the server. We assume stable communication links, where the student model parameters and intermediate representations are ideally transferred from the clients to the server without any loss.

\begin{table*}[!t]
  \centering
  \caption{Training parameters}{
    {\renewcommand{\arraystretch}{0.95}
    \begin{tabular}{c c c}
      \toprule
      Parameters & Values for CIFAR-10 & Values for CIFAR-100\\
      \midrule
      Sum of clients' data & 1000 to 20000 & 4000 to 20000 \\
      Number of pre-training epochs & \multicolumn{2}{c}{200}\\
      Number of clients to choose from in one round of FL step in FedHNP $C$ & \multicolumn{2}{c}{10}\\
      Number of client epochs within one round of FL in FedHNP $E_{\mathrm{FL}}$ & \multicolumn{2}{c}{$\lfloor2\times10^5/\mathrm{sum\ of\ clients'\ data}\rfloor$}\\
      Number of rounds of FL for FedHNP $R$ & 20 & 100\\
      Number of epochs of personalization in FedHNP $E_p$ & 10 & 20\\
      Number of HND training epochs for method 2 and 3 & $\lfloor 4\times10^6/\mathrm{sum\ of\ clients'\ data}\rfloor$ & $\lfloor 2\times10^7/\mathrm{sum\ of\ clients'\ data}\rfloor$\\
      \bottomrule
    \end{tabular}}\label{tab:trainingparams}
}
\end{table*}
\begin{table*}[!t]
  \centering
  \caption{Compared methods}{
  \resizebox{0.95\textwidth}{!}{
    {\renewcommand{\arraystretch}{0.95}
    \begin{tabular}{l c c c c c}
      \toprule
      & & 
      \begin{tabular}{c}
        Is the client data label needed? 
      \end{tabular} & 
      \begin{tabular}{c}
        Is the labeled data \\ for pre-train needed?
      \end{tabular} &
      \begin{tabular}{c}
      Is communication \\ during training needed?
      \end{tabular} & 
      \begin{tabular}{c}
        Is the privacy of \\ the client data maintained? 
      \end{tabular}\\
      \midrule
      Method1 & \textbf{FedHNP (Proposed)} & No & Yes & Yes & Yes\\
      Method2 & Centralized HND 
      & No & Yes & No & No\\
      Method3 & 
      Individual HND & No & Yes & No & Yes\\
      \bottomrule
    \end{tabular}}\label{tab:methods}
    }
}
\end{table*}
\cref{tab:trainingparams} shows the settings. The number of epochs for pre-training is the same for both methods to align the condition. For CIFAR-100, the number of epochs for local HND is higher than for CIFAR-10 because the task is more complicated and takes longer to learn. The number of training epochs increases as the clients' data volume decreases to equalize the number of gradient updates when the data volume changes. We use Adam\cite{diederik2015adam} to optimize the parameters by minimizing the loss function shown by \cref{eq:hndloss} in training client models.

\vspace{2mm}
\noindent\textbf{Compared methods.}\quad
\cref{tab:methods} shows compared methods. Both use HND, but the scope of data use and training methods differ. The proposed method, referred to as FedHNP, is compared with the conventional HND under different data constraints.

\noindent\texttt{Centralized HND:} Centralized HND assumes an ideal scenario where client data can be aggregated, meaning there are no data constraints. In this baseline, the clients' datasets are collected, combined, and used for model training with HND on the server. The goal of the proposed method is to achieve accuracy comparable to this centralized approach.

\noindent\texttt{Individual HND:} Similar to the proposed method, this approach assumes a scenario where client data cannot be aggregated, and each client performs HND individually using only its own data. Due to the reduced amount of available data during HND, accuracy is expected to deteriorate, particularly when the number of client data samples is small.

\subsection{Results}\label{subsec:result}
\subsubsection{Reduction of Computational and Communication Costs via Model Replacement}
\label{sec:cost}
\begin{table*}[!t]
  \centering
  \caption{Computation and Communication Cost Reduction via Split Computing vs. Cloud Computing. }{
    {\renewcommand{\arraystretch}{0.95}
    \begin{tabular}{c c c c c c c}
      \toprule
      \multirow{2}{*}{Methods} & 
      \multirow{2}{*}{Split point} &
      \multicolumn{2}{c}{Computation load (FLOPs)}
      &
      \multicolumn{3}{c}{Communication traffic}\\
      \cmidrule(lr){3-4}
      \cmidrule(lr){5-7}
      && Client & Server & Data shape & Data volume (8bit) & Data volume (32bit)\\
      \midrule
      \midrule
      Cloud computing using original model & - & 0 & $1.40\times10^8$ & (3, 32, 32) & 3.1 KB & 12.3 KB\\
      \midrule
      Split computing using original model & 2 &$1.77\times10^6$ & $1.38\times10^8$ &(64, 16, 16) & 16.4 KB & 65.5 KB\\
      (two split points are listed) & 8 & $5.42\times10^7$ & $8.60\times10^7$ &(128, 8, 8) & 8.20 KB & 32.8 KB\\
      \midrule
      Split computing using replacement model & 3 & $2.02\times10^6$ & $1.04\times10^8$ &(12, 9, 9) & 1.0 KB & 3.9 KB\\
      \bottomrule
    \end{tabular}}\label{tab:model_comp}}
\end{table*}

We begin by demonstrating the impact of replacing the original model with the trained Head and Tail networks on both computational and communication loads. \cref{tab:model_comp} outlines the computational load on the client and server, along with the communication traffic generated during inference when using cloud computing, SC with the original model, and SC with the model trained by the proposed method. The split point refers to the layer at which the model is divided, starting from the input layer. Computational load is measured by the number of floating-point operations (additions and multiplications) performed by the models. Data volume corresponds to the size of the intermediate representation transmitted from the client during inference, presented in both the commonly used 32-bit floating-point format and an example of 8-bit quantization-based compression. For cloud computing, communication traffic is calculated under the assumption that input data (e.g., images) are directly transmitted.

As shown in \cref{tab:model_comp}, replacing the original model with the proposed Head and Tail networks significantly improves the trade-off between computational load and communication traffic. When splitting the original ResNet-18 model, the viable split points, considering client computational load and avoiding skip connections, are at the 2nd and 8th layers. However, splitting at the 2nd layer results in high communication traffic, while splitting at the 8th layer imposes excessive computational load on the client, rendering both points unsuitable for split computing in their original form.

In contrast, by replacing the original model with the one trained using the proposed method, the overall computational load is reduced by 24\% compared to the original model, with client-side computation kept at a similar level to the 2nd-layer split. Furthermore, communication traffic is drastically reduced—down to one-third of that in cloud computing and 1/8 to 1/16 of that in split computing with the original model. These results highlight the effectiveness of replacing the Head and Tail networks when implementing split computing.

\subsubsection{Comparison of Training Curves Across Different Training Methods}\label{subsubsec:basecomp}
\begin{figure}[!t]
  \centering
  \includegraphics[width=0.45\textwidth]{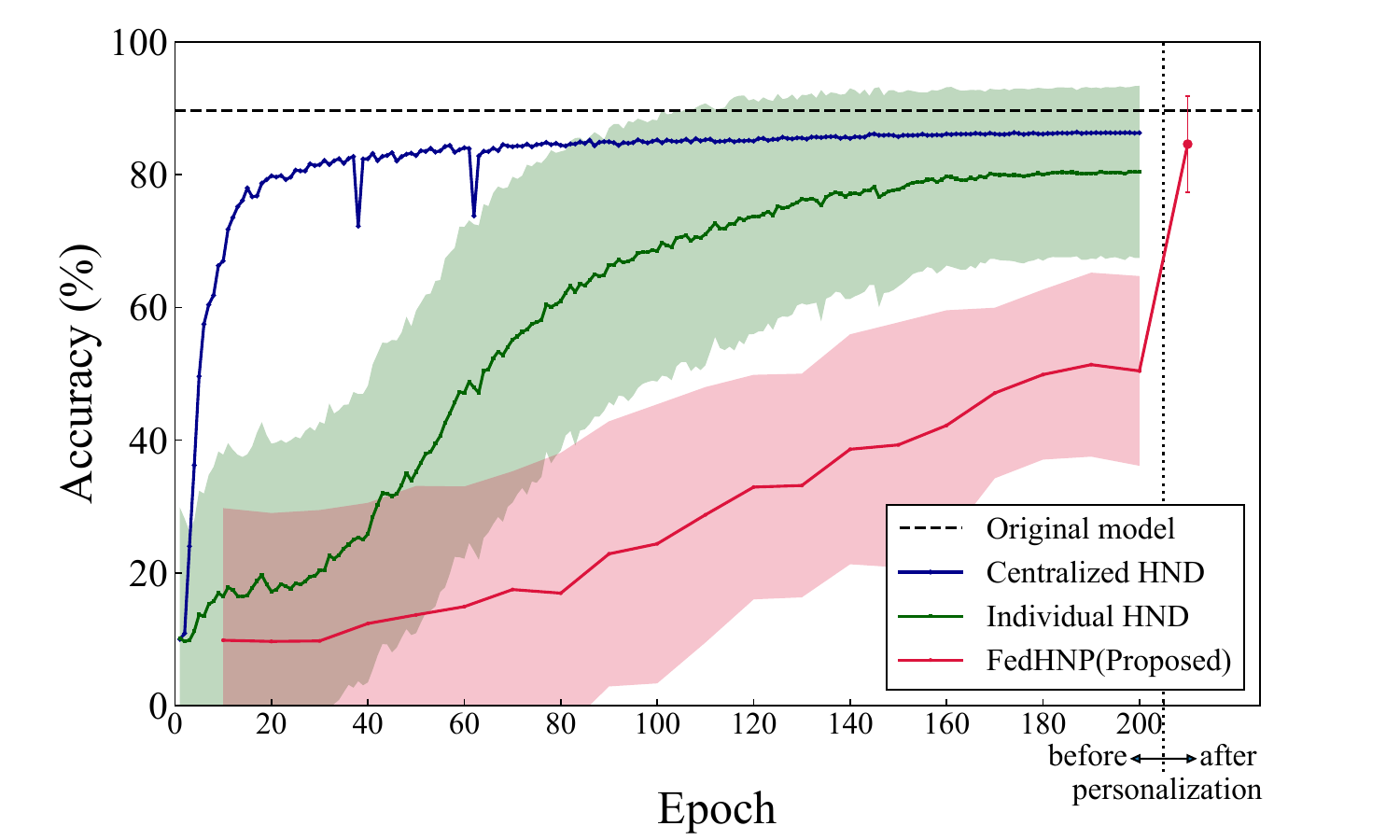}
  \caption{Comparison of training curves with CIFAR-10, 20000 client training samples. For FedHNP, we plot the stage 1 accuracy for each of the ten epochs and the accuracy after stage 2 in the ``personalized" position.}\label{fig:acc_comp}
\end{figure}

As part of the evaluation of model accuracy for each method, we first present the training curves versus training epochs for the proposed method and the baseline methods. For the proposed method, FedHNP, since training occurs in rounds of ten epochs, we report the test accuracy after each round during the FL stage, as well as the final accuracy after personalization.

\cref{fig:acc_comp} illustrates the test accuracy at each training epoch using CIFAR-10 with 20,000 client training samples. The dashed line represents the accuracy of the original model trained with 30,000 pre-training samples. For both FedHNP and Individual HND, the models trained on each client are evaluated using the test dataset assigned to that client, with the average and standard deviation of the results depicted. In contrast, for the Original model and Centralized HND, evaluation is performed using the aggregated test data from all clients.

In \cref{fig:acc_comp}, Centralized HND, which trains the student model under ideal conditions, shows the fastest increase in accuracy, with final accuracy approaching that of the original model. In contrast, Individual HND demonstrates a slower accuracy increase, ultimately converging at a lower level.

Meanwhile, FedHNP exhibits the slowest accuracy increase during the FL-based training stage, achieving significantly lower accuracy after 200 epochs. However, client personalization significantly boosts model performance, with the mean accuracy becoming comparable to or even exceeding that of the original model while still preserving client data privacy. This improvement is likely because, while Centralized HND trains a single model capable of recognizing all classes, FedHNP leverages personalization to tailor the model to each client’s training dataset, which is consistent with the distribution of that client’s test dataset. On the other hand, although Individual HND is trained on a dataset with a distribution consistent with the test dataset, it does not achieve the same high performance as FedHNP.

These results suggest that while the Federated Learning stage in FedHNP does not directly yield high accuracy on each client’s test data, it serves as an effective pre-training mechanism for the student model. This pre-training significantly enhances the personalization stage, enabling efficient fine-tuning of the model.

\subsubsection{Accuracy Evaluation Under Different Dataset Configurations}\label{subsubsec:result1}
\begin{figure}[!t]
  \centering
  \begin{minipage}[b]{0.49\columnwidth}
    \centering
    \includegraphics[width=1.0\columnwidth]{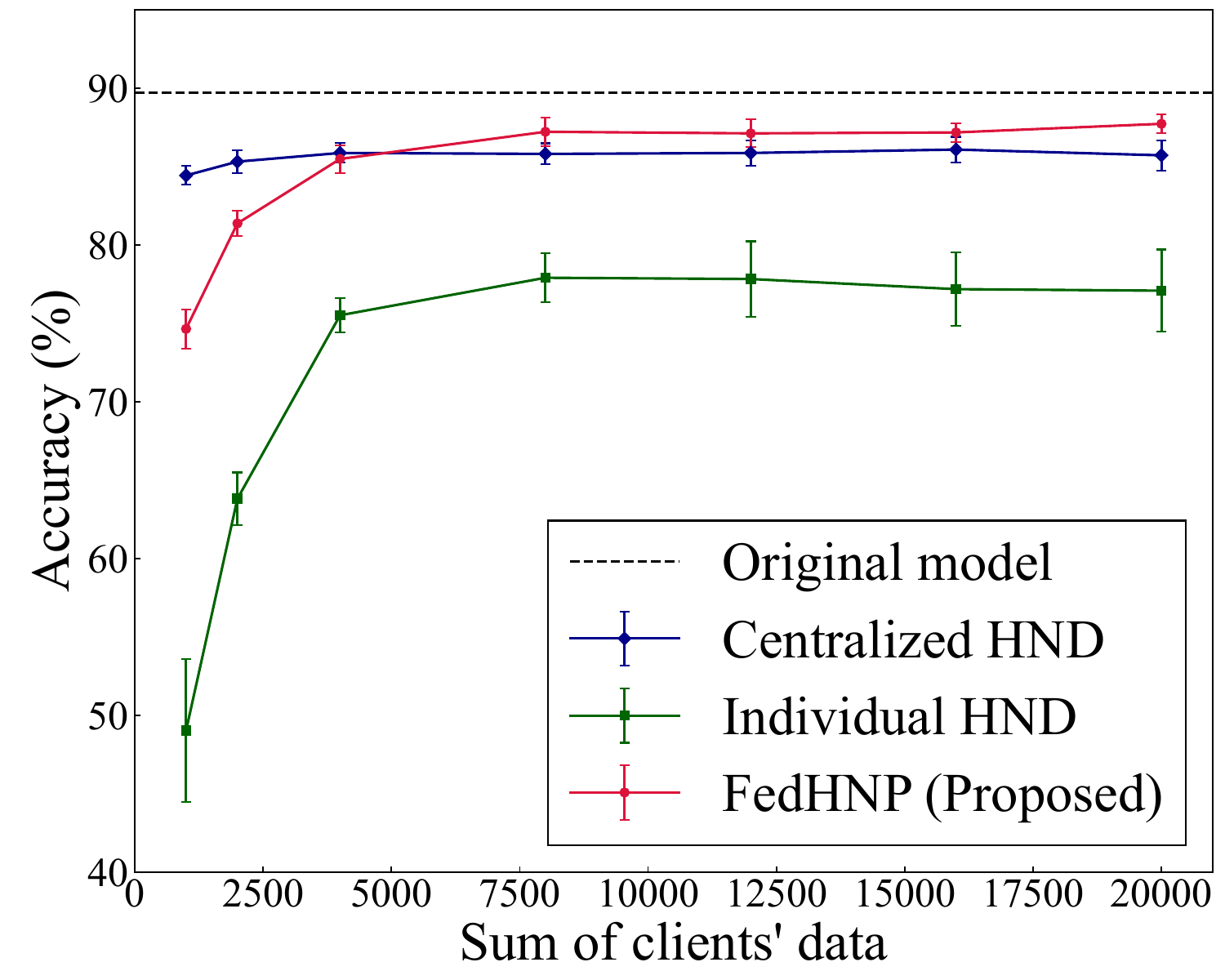}
    \subcaption{CIFAR-10}
    \label{subfig:acc_cifar10}
  \end{minipage}
  \begin{minipage}[b]{0.49\columnwidth}
    \centering
    \includegraphics[width=1.0\columnwidth]{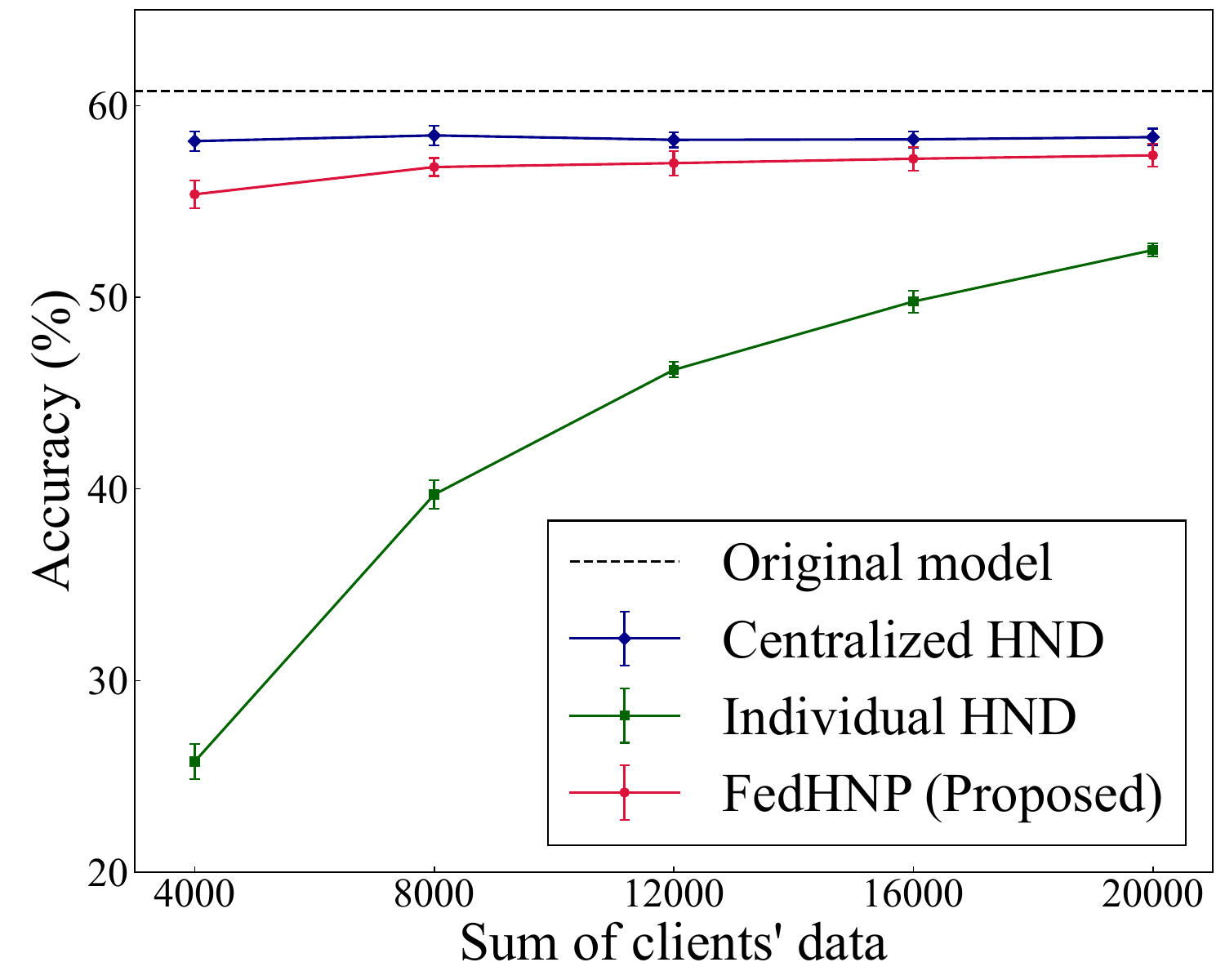}
    \subcaption{CIFAR-100}
    \label{subfig:acc_cifar100}
  \end{minipage}
  \caption{CIFAR-10 / CIFAR-100 test accuracy in different total number of clients data}\label{fig:differentdata_acc}
\end{figure}

\cref{fig:differentdata_acc} shows the test accuracy for each method using the CIFAR-10 and CIFAR-100 datasets, with varying numbers of client training samples. For CIFAR-10, we use between 1,000 and 20,000 total samples, corresponding to 20 to 400 samples per client. For CIFAR-100, due to the difficulty of training with too few samples, we use between 4,000 and 20,000 total samples, corresponding to 80 to 400 samples per client. The dashed line indicates the accuracy of the original model, which is pre-trained and thus independent of the total client data. Additionally, for the proposed method, the accuracy after personalization is shown. The results represent the average of 10 trials, along with their standard deviation, each conducted with a different random seed.

As the number of client data samples increases, the accuracy of all methods improves, though none surpass the accuracy of the original model. This is due to the fact that, as shown in \cref{sec:cost}, the head network replaces parts of the original model with a smaller number of model parameters and a bottleneck to reduce computational and communication costs. Despite these challenges, the proposed method still achieved model accuracy comparable to that of Centralized HND, which operates under ideal conditions. Even in cases where client datasets were small, although the proposed method lagged behind the original model by a dozen points and fell a few points short of Centralized HND, it still outperformed Individual HND by several dozen points. This indicates that the application of the Federated Learning framework significantly mitigates the accuracy degradation caused by insufficient client data.

We note that the non-IID setting used in this evaluation, in which each client holds only 2 out of 10 classes (CIFAR-10) or 10 out of 100 classes (CIFAR-100), corresponds to an extreme label-distribution skew, often referred to as a pathological non-IID setting~\cite{mcmahan2017fa}. In addition, the sampling ratio $\alpha_{i,c}\sim U(0.4, 0.6)$ introduces quantity skew across clients. The reported results therefore characterize the behavior of FedHNP under one of the severest heterogeneity conditions. Qualitatively, as the heterogeneity decreases toward IID, the Stage-1 global model approaches the behavior of Centralized HND and the additional gain of personalization diminishes; that is, FedHNP is expected to degrade gracefully to its FL backbone under mild heterogeneity. A systematic sweep of intermediate heterogeneity levels, e.g., using Dirichlet-based partitions, requires re-running the full training pipeline and is left as future work.

\subsubsection{Evaluation of Personalization Effectiveness in FedHNP}\label{subsubsec:personalizationeffect}
In the results presented so far, the proposed FedHNP, by incorporating HND, enabled local model training with unlabeled data. The use of the FL framework primarily acted as an effective pre-training mechanism for the student model. Moreover, personalization significantly improved accuracy to levels comparable to the original model. Here, we delve deeper into the impact of personalization in FedHNP. 

\begin{figure}[!t]
  \centering
  \begin{minipage}[b]{0.49\columnwidth}
    \centering
    \includegraphics[width=1.0\columnwidth]{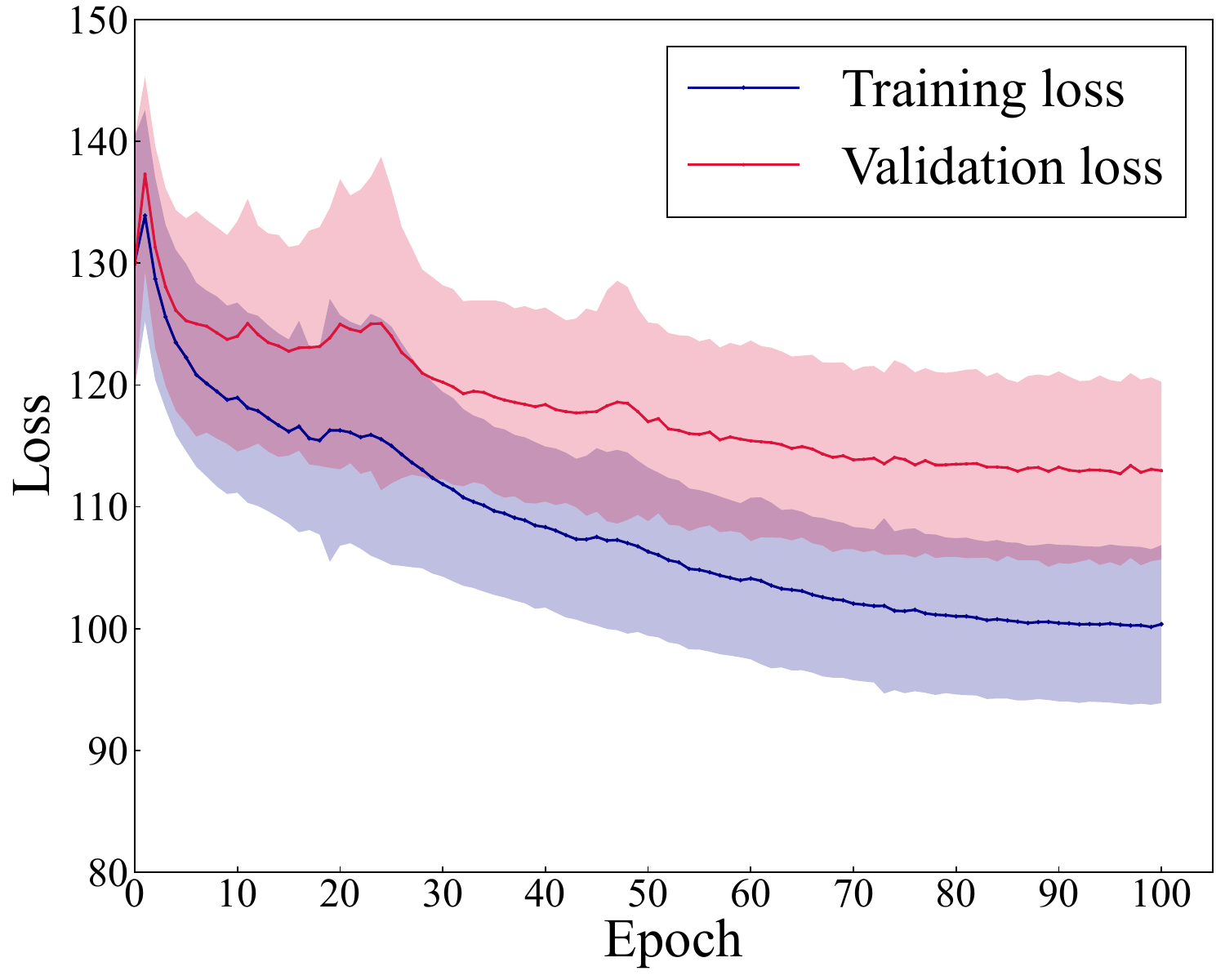}
    \subcaption{Loss}
    \label{subfig:phase2loss}
  \end{minipage}
  \begin{minipage}[b]{0.49\columnwidth}
    \centering
    \includegraphics[width=1.0\columnwidth]{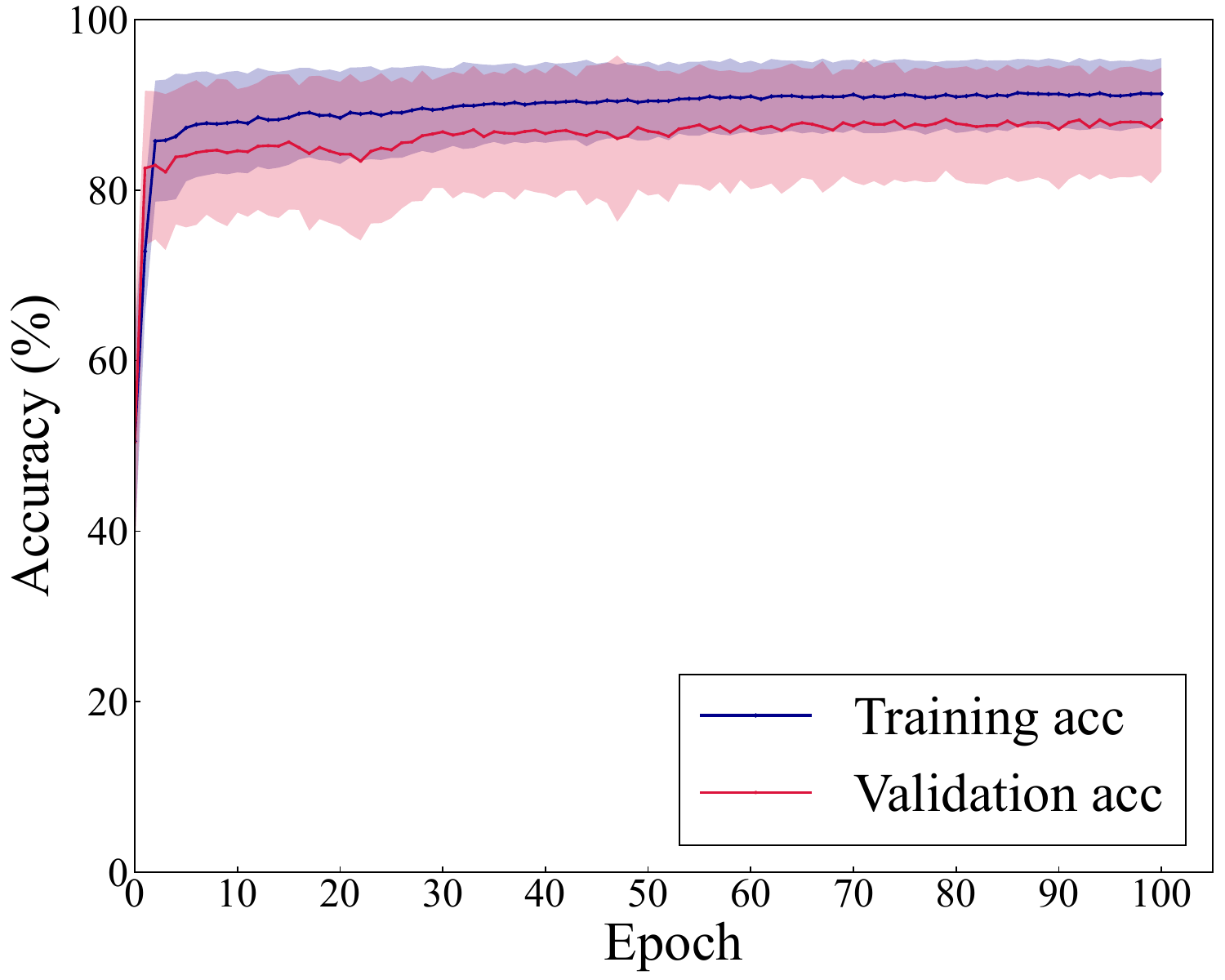}
    \subcaption{Accuracy}
    \label{subfig:phase2acc}
  \end{minipage}
  \caption{Stage 2 training curve with CIFAR-10, 20000 data. Personalizing for 100 epochs.}\label{fig:phase2}
\end{figure}

\cref{fig:phase2} illustrates the training and validation loss, as well as accuracy, plotted against training epochs during the client personalization stage. The dataset is CIFAR-10 with 20,000 client samples. The loss and accuracy at epoch 0 represent the values from the global model trained in stage 1.
As shown in \cref{fig:phase2}, both training and validation losses drop sharply in the first few epochs, followed by a gradual decline. Similarly, accuracy improves substantially during the initial fine-tuning epochs, then levels off with a slower rate of improvement. This indicates that even the relatively small number of fine-tuning epochs (10 epochs) used in the previous experiments was sufficient to significantly enhance accuracy for this task. Furthermore, it suggests that dedicating additional computational resources to personalization could further enhance the model's accuracy.

\begin{figure}[!t]
  \centering
  \begin{minipage}[b]{0.49\columnwidth}
    \centering
    \includegraphics[width=1.0\columnwidth]{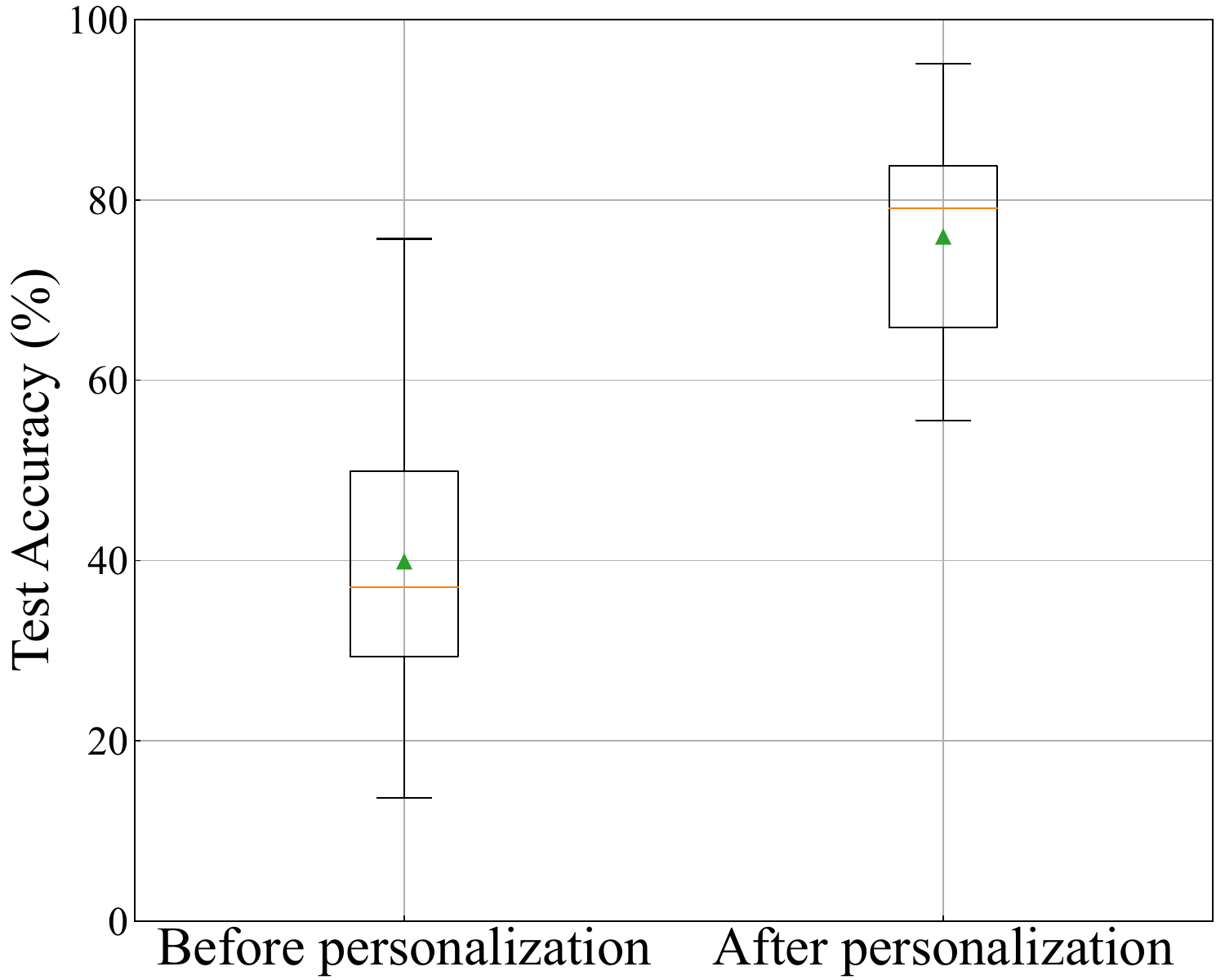}
    \subcaption{Case1: 1000 data for clients}
    \label{subfig:box1}
  \end{minipage}
  \begin{minipage}[b]{0.49\columnwidth}
    \centering
    \includegraphics[width=1.0\columnwidth]{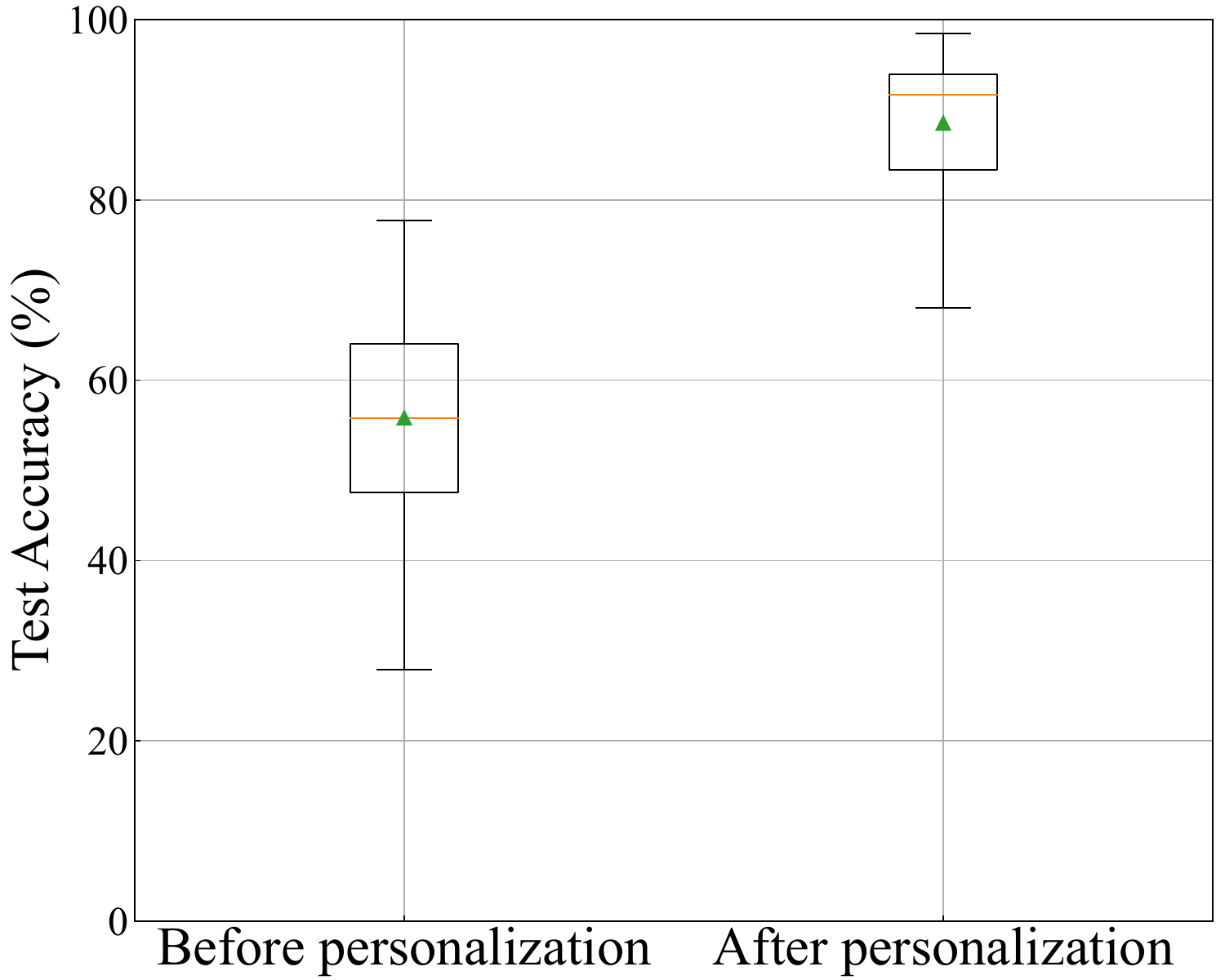}
    \subcaption{Case2: 20000 data for clients}
    \label{subfig:box2}
  \end{minipage}
  \caption{Test accuracy before / after personalization in FedHNP. The triangle represents the mean, and the line in the box represents the median.}\label{fig:bf_af_personalization}
\end{figure}

\cref{fig:bf_af_personalization} presents box-and-whisker plots of client model test accuracy before and after personalization, using 1,000 and 20,000 client data samples. The orange line represents the median, while the green triangle denotes the mean accuracy across clients. The box reflects the interquartile range (IQR), capturing the middle 50\% of the data, while the whiskers extend to 1.5 times the IQR, showing the overall data spread.

Personalization not only significantly boosts accuracy but also reduces the variation in model accuracy between clients. This effect is particularly pronounced when the number of client data samples is small, with the median accuracy increasing by 40 points. Additionally, in cases where clients have 20,000 data samples, more than 50\% of clients achieve accuracy levels of 90\% or higher, comparable to the original model.

Thus, personalization is a crucial component in the proposed FedHNP. The integration of HND, FL, and personalization proves to be highly effective, allowing for the training of accurate Head Networks using only small amounts of unlabeled data.

\subsubsection{Discussion: Reliance on the Pre-trained Model}\label{subsubsec:pretrained_discussion}
Finally, we discuss the reliance on the pre-trained model. Because HND trains the student to reproduce the intermediate representations of the teacher extracted from the pre-trained model, the accuracy of the pre-trained model acts as an effective upper bound for the student: the dashed lines in \cref{fig:differentdata_acc} represent this empirical ceiling, and none of the distillation-based methods exceed it. Conversely, if the pre-trained model is poorly trained for the target task, FedHNP inherits its deficiency. We consider the availability of a well-trained model a mild assumption in practice, since high-quality pre-trained models for common tasks are widely available in public repositories, whereas their original training datasets are typically not released---which is precisely the situation our method targets. When no suitable pre-trained model exists, supervised training of the full model is unavoidable and our method is not applicable.

%% file: 5_conclusion.tex
\section{Conclusion}\label{sec:conclusion}
In this study, we proposed an unsupervised training method for Head Networks that reduces both computational load on mobile devices and communication with the server, even under highly data-constrained conditions. The method integrates HND, a distillation-based training approach, with the FL framework, a collaborative model training method, to address challenges posed by unlabeled and limited training data. Additionally, by introducing a personalization technique, the drop in inference accuracy resulting from model replacement with the Head Network is effectively mitigated.
Experimental evaluations demonstrate that, despite the stringent data constraints unique to our proposed method, it achieves accuracy comparable to baselines with fewer or no data constraints. The Head Network trained by our method reduces communication requirements on mobile devices by 95\%, with only a 3-5 point drop in accuracy compared to the original model. Additionally, the results show that executing personalization—specifically fine-tuning—on the client side can increase inference accuracy by more than 30 points, further highlighting the significant role personalization plays in mitigating accuracy degradation when using compact Head Networks.

Future work will focus on developing methods to determine the optimal model architecture for each device. In this study, we assumed a fixed model structure across all clients. However, in real-world applications, computational power and communication delays are likely to vary between devices and over time. Therefore, it is essential to explore approaches that tailor model architectures to individual devices or prepare multiple models and select the most appropriate one for each device.